\documentclass{article}
\usepackage{ijcai22}

\usepackage{times}
\usepackage{soul}
\usepackage{url}
\usepackage[hidelinks]{hyperref}
\usepackage[utf8]{inputenc}
\usepackage[small]{caption}
\usepackage{graphicx}
\usepackage{amsmath}
\usepackage{amsthm}
\usepackage{booktabs}
\usepackage{algorithm}
\usepackage{algorithmic}
\title{ConjointNet: Enhancing Conjoint Analysis for Preference Prediction \\with Representation Learning}

\author{
    Yanxia Zhang \and Francine Chen \and Shabnam Hakimi \and Totte Harinen \and Alex Filipowicz \and \\ 
    Yan-Ying Chen \and Rumen Iliev \and Nikos Arechiga \and Kalani Murakami \and \\ 
    Kent Lyons \and Charlene Wu \and Matt Klenk \\
    \affiliations
    Toyota Research Institute
    \emails
    yanxia.zhang, francine.chen, shabnam.hakimi, totte.harinen, alex.filipowicz, \\ 
    yan-ying.chen, rumen.iliev, nikos.arechiga, kalani.murakami.ctr, \\
    kent.lyons, charlene.wu, matt.klenk@tri.global
}

\begin{document}

\maketitle

\begin{abstract}
  Understanding consumer preferences is essential to product design and predicting market response to these new products. Choice-based conjoint analysis is widely used to model user preferences using their choices in surveys. However, traditional conjoint estimation techniques assume simple linear models. This assumption may lead to limited predictability and inaccurate estimation of product attribute contributions, especially on data that has underlying non-linear relationships. In this work, we employ representation learning to efficiently alleviate this issue. We propose ConjointNet, which is composed of two novel neural architectures, to predict user preferences. We demonstrate that the proposed ConjointNet models outperform traditional conjoint estimate techniques on two preference datasets by over 5\%, and offer insights into non-linear feature interactions. 
\end{abstract}

\section{Introduction}
Knowledge of consumer preferences is central not only to designing product features but also to predicting market response to new products or services. Conjoint analysis is frequently used to model consumers' choices, thereby gaining insight into their preferences for specific products or product attributes \cite{green2001thirty}. Conjoint analysis helps businesses identify the best attributes to include in a product. Choice-based (or discrete choice) conjoint analysis is perhaps the most common conjoint analysis approach, leveraging consumers' actual choices rather than more abstract ratings or rankings to infer preference over various product attributes. Respondents make choices over many combinations of product attributes, signaling their preference through their behavior. The conjoint analysis method effectively detects these signals, learning the impact of each product attribute on users’ choices and providing preference estimates that can be used to predict real-world decisions across multiple domains, ranging from purchase of consumer products to adoption of clinical interventions \cite{green2001thirty,orme1997assessing}. 

Everyday choices illustrate how choice-based conjoint analysis captures preferences over consumers' often large set of options. Take car purchasing as an example. If a car dealer wants to know consumers' preferences over their inventory of sedans, they can create a choice scenario that asks consumers to choose configured sedans at different prices. The car options might be a combination of \textit{attributes} (e.g., brand, size, color, price), which are available in different \textit{levels} (e.g., the size may be 2-door or 4-door). Choice-based conjoint analysis identifies how the different attributes, i.e their \textit{partworths}, influence consumers choices. This approach is also often used to estimate consumers' \textit{willingness to pay} for specific product attributes. However, this method is limited to testing explicit attributes (e.g., size, color) and cannot learn user preferences for implicit style features (e.g., cuteness, comfortableness). Further, as the number of attributes grows, the cost to both the survey designer and respondents also increases, making it impractical to collect data that truly reflects the complexity of the choice space.

Recent advances in Deep Neural Networks (DNNs) present new opportunities for addressing these limitations. Specifically, representation learning is an emerging type of approach that allows a model to automatically discover features from data by training a neural network \cite{Bengio2013-yx}. Representation learning techniques have shown great success and revolutionized multiple fields including computer vision \cite{He2015-ms}, text analysis \cite{Lai2015-qt}, Natural Language Processing (NLP) \cite{Devlin2018-ck,radford2018improving} as well as speech recognition \cite{amodei2016deep}. In this type of method, neural network architectures are designed to encode the raw data (i.e., images, text or audio signals) into intermediate representations, which will then be used to speed up various downstream tasks such as image recognition or text classification.

Our motivation for applying representation learning to conjoint analysis is three-fold. First, we aim to improve the efficiency of processing survey data with an end-to-end training system. In this way, feature learning and prediction are optimized simultaneously. Unlike prior works that applied Support Vector Machine and Hierarchical Bayesian Modeling on choice-based conjoint surveys \cite{toubia2007optimization,chapelle2005machine}, our approach eliminates the need for feature engineering and is therefore able to scale up easily on a large number of input features. For example, the Moral Machines dataset \cite{Awad2018-zc} has an input size of over 20 attributes with as many as 5 levels per attribute. The number of interactions grow exponentially as the input parameters increase. A brute force permutation would generate over 11,000 three-way interactions and it is impractical to test all possible hypotheses. Second, our approach learns non-linear features from data. This helps a domain expert to use multi-way interactions that have not been considered before or consider new implicit features like "cuteness". Lastly, the use of representation learning neural network architectures can enable new possibilities of learning preference from multimodal data beyond surveys and data where human responses are sparse. 

Our contributions are: 
\begin{itemize}
    \item ConjointNet enhances conjoint analysis by discovering non-linear interactions from data. This complements existing workflow that requires a domain expert to design new interaction features. Our results demonstrate significantly improvement in predictive performance over traditional conjoint analysis on two public datasets. 
    \item ConjointNet enables end-to-end learning and works with raw choice-based conjoint survey data without hand-crafted features. This provides the flexibility of working with different target responses from a same set of survey input. The resulting representations can be easily concatenated with other modalities such as images or personal embeddings. 
    \item ConjointNet employs two novel architectures that are designed to effectively predict user preferences over unseen data in addition to partworth estimation. The semi-supervised ConjointNet employs auto-encoders to pretrain on raw inputs. This provides the benefit of leveraging a large amount of unlabelled data, and therefore requires fewer observations per respondent. The residual ConjointNet uses a ResNet-inspired \cite{He2015-ms} architecture to simultaneously learn the linear and non-linear components.
\end{itemize}

\section{Related Work}

\subsection{Discrete Choice Models}
Traditional models used for choice-based conjoint analysis weight the impact of different attributes on people's choices. Common methods include logistic or multinomial regression models that estimate the weights of different product attributes, which are then summed to estimate the utility of different choice options. \cite{steiner2018user}. These types of linear additive models can be particularly successful in circumstances where attributes are independent of and do not interact with other attributes \cite{orme2006getting}. In cases where attributes are correlated, methods such as ridge regression or hierarchical Bayesian modeling provide regularization that can help improve the precision of the weight estimates \cite{toubia2007optimization,chapelle2005machine,steiner2018user}). However, although useful to address attribute correlations, these approaches lack the sensitivity to characterize latent attribute spaces, missing an opportunity to better understand how clusters of attributes work together to influence choices. Addressing interactions between attributes are more challenging, as these generally need to be specified \emph{a priori} and can be computationally expensive to discover as the number of attributes grows. 

In addition to challenges with correlated and interacting attributes, a number of decision characteristics follow non-linear decision rules (e.g., non-compensatory decision rules \cite{steiner2016customer}). Although these non-linear decision rules can be modeled, they must again be specified \emph{a priori} to be modeled appropriately \cite{steiner2018user}. To address these limitations with traditional choice-based conjoint analysis techniques, more recent techniques have focused on using machine learning tools to gain a more comprehensive understanding of the relationships between attributes and product choices. We highlight the advantages and shortfalls of these methods below.

\subsection{Feature Selection}
Commonly used feature selection approaches include Lasso \cite{Tibshirani1996-id} and Bayesian variable selection \cite{o2009review} to choose a subset of features that are based on their importance for prediction. Tree-based learning is another widely used type of techniques to pick features with the most information gain. Random forests grow a large number decision trees with a subset of randomly selected features \cite{Ho}. Another greedy approach is based on Maximum Relevance Minumum Redudancy (MRMR) which tries to select non-redundant features, e.g. \cite{unler2011mr2pso}. These approaches start with a set of global features, and cannot discover new features that do not exist in the original set. 

\subsection{Representation Learning}
An autoencoder (AE) learns embeddings (a lower-dimensional representation) from data without labels. It is usually comprised of two parts: an encoder that transforms the input data to a latent space (usually at a lower dimensions) and a decoder that reconstructs the input from the latent space. The learned embeddings are used to improve the performance of downstream tasks \cite{Bengio2013-yx}. While it is easy to obtain good reconstruction results with an ordinary AE, efforts have been made to regularize the latent space in AE and prevent overfitting. A Variational  AutoEncoder (VAE) \cite{Kingma2013-wu} regularizes the latent space by enforcing the latent variable to be a normal distribution, and can generate data by sampling the latent space. There are a few works that applied deep learning for preference modeling. Loreggia et al. \cite{loreggia2019cpm} proposed a Siamese networks for learning a metric (distance) between set of objects to represent the preferences of two users. Pfannschmidt et al. \cite{pfannschmidt2022learning} applied neural network to learn generalized utility functions that are context-dependent.

\begin{figure*}[htp]
    \centering
    \includegraphics[width=.9\linewidth]{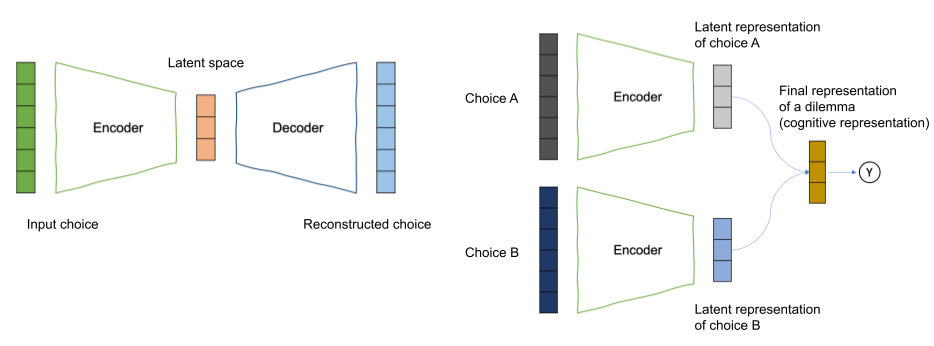}
    \caption{Proposed SSL ConjointNet Architecture on Choice Problems}
    \label{fig:ssl}
\end{figure*}

\section{Choice-based Conjoint Analysis}
Since its inception in the 1970s, the literature on conjoint analysis methods and research designs has become rich and varied \cite{green2001thirty}. In one of the simplest versions of the method, $n$ consumers are asked to choose between multiple options in a number of trials. The responses are training samples $(x, y)$ where $x$ represents the item and $y$ is the target. Each item is described by $m$ attributes with a total $k_{i}$ levels for each attribute $i=1, ..., m$. In choice-based surveys, the target $y$ is the observed choice variable (ground truth label) and is equal to one if the choice item is selected and zero otherwise. User preferences are modeled as a utility function $U(x)$ which represents how much users value the product. 

Thus, choice-based conjoint analysis is essentially a linear model that assumes the utilities of an item $U(x) = wx$ are the sum of the partworth values of all attributes defined as:
\begin{equation}
    U(x) = \sum_{i=1}^{m}\sum_{j=1}^{k_{i}}w_{ij}x_{ij}
    \label{equ:conjoint}
\end{equation}
where the partworth value $w_{ij}$ represents the utility of the $i$th attribute at level j \cite{green2001thirty}. 

We can rank users' preference over each attribute at different levels using the partworth value $w_{ij}$. Additionally, the importance of each attribute is determined by either summing up the partworths at all levels or defined as $u_{i} = max(w_{ij}) - min(w_{ij})$. Finally, an option is selected by using attribute levels with maximum partworth sums.

\section{ConjointNet: Conjoint Analysis with Representation Learning}
Although linear regressions for conjoint analysis as shown in Equation \ref{equ:conjoint} are successful when all input features are independent, this is hard to get in real-world datasets. Certain features are inherently correlated, for instance, "engine capacity and fuel" consumed in cars. Another way to overcome the limitation of linear assumption is to use data collected from a completely randomized design experiment. However, this is often not the case in real-world observational data. Instead of relying on the assumption that attributes are independent, ConjointNet allows approximation with non-linear neural networks and thus can model non-linear feature interactions.

\subsection{ConjointNet with Semi-supervised Learning}
Our first design of the ConjointNet architecture is based on semi-supervised learning. This network makes use of both labelled and unlabelled data samples. Fig. \ref{fig:ssl} shows the architecture that employs an antoencoder (left) to transform the input choice data into a latent representation, and a classification network (right) that makes predictions from learned representations extracted by the autoencoder. 

\subsubsection{Autoencoders for Representation Learning}
An autoencoder is a building block for deep learning as a feature learning technique that maps raw inputs into a latent space. It coverts an unsupervised problem to a supervised problem by reconstructing the original inputs with a loss function. We implement an autoencoder that comprises 3 hidden layers of neurons for both the encoder and decoder (see Fig. \ref{fig:ssl} left). The autoencoder is symmetric with the size of the input layers matching that of the output layer. The latent representation is one bottleneck layer in the middle. Given the input items $X = {x_{ij}}$, where $i\in{[1,m]}, j\in{[1,k]}$, the optimization function of the autoencoder is defined as:
\begin{equation}
    L_{recon} = min \sum_{i=1}^{m}\sum_{j=1}^{k}D(x_{ij}, \tilde{x}_{ij})
\end{equation}
where $D$ is a distance function such as $|x-\tilde{x}|$ where $\tilde{x}$ is the reconstructed choice. The new representation $h_{ij} = g(Wx_{ij}+b)$ is transformed from raw inputs $x_{ij}$. Then it can be used to reconstruct output $\tilde{x}_{ij} = f(W^\mathsf{T} h_{ij} + b^{\prime})$. Weights $W$, biases $b$ and $b^{\prime}$ are learned through back propagation. Compared to PCA, autoencoders are more powerful and can learn non-linear representations because of the non-linear activation functions $f$ and $g$. In this work, we implemented two variants of autoencoders, namely, the ordinary AE and VAE. Because there are no ordinal relationship between different attribute categories and levels. The input $x$ is given as a categorical variable. Each attribute is converted with a one-hot encoding, and can be concatenated as either a 1D or 2D vector (one attribute per row) before feeding to AE.  

\subsubsection{Choice Classification Network}
Fig. \ref{fig:ssl} illustrates an example of the architecture of the choice classification network that predicts user choices over two items. Given a pair of input items $(x_{A}, x_{B})$, we first pretrained the AE without choice labels and then used the encoder to obtain the latent vector for choices A and B, denoted as $h_{A}$ and $h_{B}$ respectively. The embeddings for choices A and B are then concatenated and fed into a multi-layer neural architecture. The final output layer is the predicted utility score $\tilde{U}(x) = \phi(h_{A}, h_{B})$. The training is optimized by minimizing the binary cross entropy between the target $\tilde{y}$ and predicted choice $\tilde{Y}$. We can easily accommodate multiple options by extending the dimension of the input layer to take embeddings from more than two choices.


\subsection{Residual ConjointNet}
The second design innovation of ConjointNet is inspired by the neural network architecture of Residual Network (ResNet) \cite{He2015-ms}. The core idea of ResNet is in the introduction of the 'identity shortcut', called a residual block, to approximate a residual function. The input $x$ is passed directly to the output of the residual block. For conjoint analysis, we are not interested in learning the residual functions with the original input $x$, rather the residual functions with the utility $U(x)$.

\begin{figure}[htp]
    \centering
    \includegraphics[width=.7\linewidth]{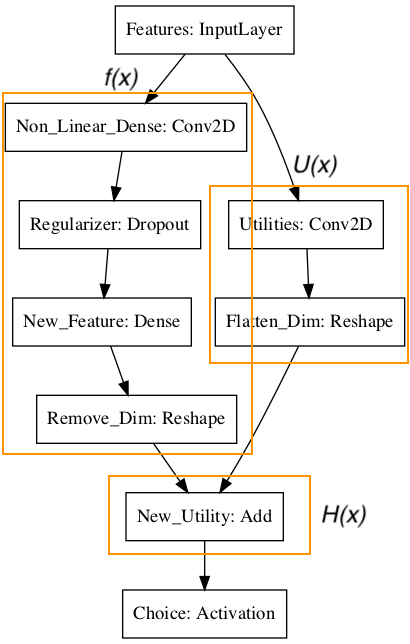}
    \caption{The proposed Residual ConjointNet jointly enforce linear and non-linear components.}
    \label{fig:resNet}
\end{figure}

Let's assume $H(x)$ represents the underlying mapping function from input $x$, we formulate our problem of learning the non-linear feature interactions:
\begin{equation}
    H(x) = U(x) + f(x)
    \label{equ:resNetconjoint}
\end{equation}
where U(x) is defined as in Eqn. \ref{equ:conjoint}. As shown in Fig. \ref{fig:resNet} the architecture consists of one hidden fully connected layer $Non_Linear_Dense$ to learn the non-linear interactions $f(x)$ which connects to the input layer. The final utility function obtained by adding the utilities extracted from new non-linear features $f(x)$ to utilities obtained from the input $U(x)$. 

\section{Experiments}
We evaluated ConjointNet on two public benchmarks, the Moral Machine~\cite{Awad2018-zc} and Car Preference~\cite{Abbasnejad2013-jx}, which were analyzed using the traditional conjoint model. One of the major differences between these two datasets is that the Moral Machine experiment did not follow a fully randomized design which is required in applying conjoint analysis. 

\subsection{Data Preprocessing}
\subsubsection{The Moral Machine (MM) Dataset}
The \emph{MM} data data is collected from a crowdsourced study that asks human subjects about moral dilemmas using a autonomous vehicles in a variation of the Trolley Problem. Subjects are presented with dilemmas in which they need to direct a self-driving car to either the left or the right side of a road.

Following \cite{Agrawal2020-cy}, we sampled only the pedestrian v.s. pedestrian dilemmas from the MM dataset, where $PedPed$ equals to '1'. Dilemmas with an empty UserID and only one respondent choice are removed from the data. After preprocessing, the final dataset size N = 15,224,624 is slightly less than the size N = 15,226,477 reported in \cite{Agrawal2020-cy}. To reconstruct the pairwise comparison for one dilemma presented to the user, we split the dataset into two sets where intervention occurs (suffix=$int$) or not (suffix=$noint$). The scenario fields which vary within pairs of responses are concatenated into one vector on index $ResponseID$.
\begin{itemize}
    \item Input features: The 42 input variables include 20 for agents on the intervention side, 20 for agents on the no intervention side, $CrossingSignal$ and $LeftHand$ which are symmetrical on both sides. We convert all variables to numeric types including categorical variables $CrossingSignal \in{\{0,1,2\}}$ and $LeftHand \in{\{0,1\}}$. An alternative would be to use one-hot encoding to represent the categorical variables, which will increase the input dimensions.
    \item Target: Given two choices for each scenario, we created the target variable $Intervened$ using the $Saved\_int$ column, represented as $Y\in\{0,1\}$. $Y=1$ indicates a choice to intervene (swerve) that leads to one set of characters being saved over the other.
\end{itemize}

\subsubsection{Car Preference Dataset}
This dataset is collected through two experiments (with 10 and 20 cars) set up in Amazon Mechanical Turk to collect pair-wise preferences. In both experiments, users were presented with a choice between two cars with different attributes. The data include input user attributes (Education, Age, Gender, Region), car attributes (Body type, Transmission, Engine capacity, Fuel consumed, Engine/Transmission layout only presented in the second experiment) and the binary target response indicating users' preferences over items. The first experiment collected data from 60 users with choices over all 45 possible configurations of attributes for 10 cars. The second experiment included 20 cars and subsets of 38 attribute combinations for each user.



\begin{figure*}[htp]
    \centering
    \includegraphics[width=.9\linewidth]{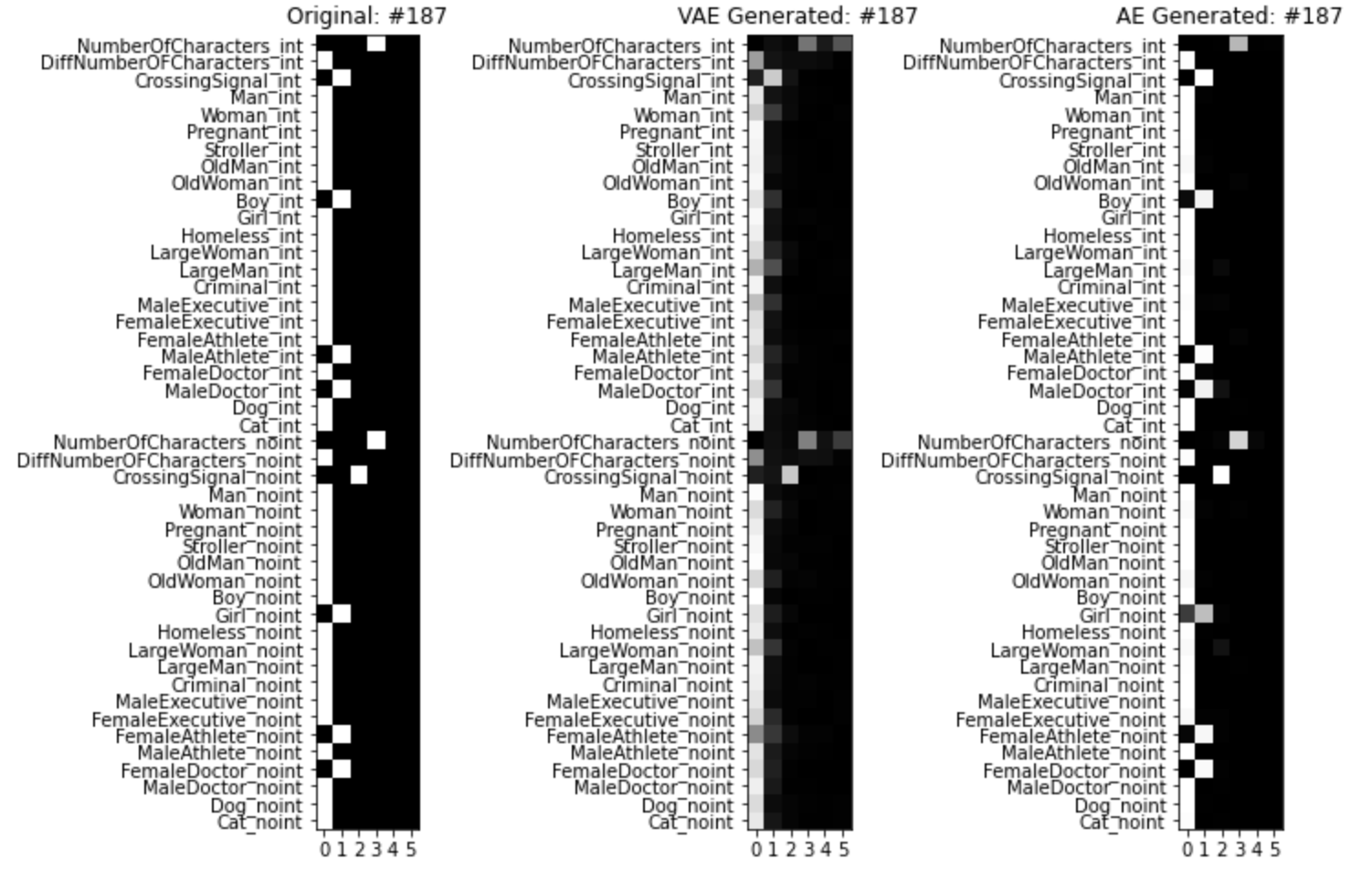}
    \caption{We show the reconstruction results of an unseen sample (left) from the MM dataset with a VAE (center) and a plain AE (right). The attributes are converted with a one-hot encoding where the corresponding level between 0 and 5 is denoted as 1 otherwise 0.}
    \label{fig:mm_recon}
\end{figure*}

\subsection{Evaluation Metrics}
We performed experiments to evaluate the performance of the different models on two public benchmarks. All train/test split ratios are roughly 70/30. For our experiments, all models converged in less than 100 epochs. The model used for testing is the one with best validation accuracy during training. We reported our results with two commonly used metrics for evaluating binary decisions: accuracy and area under the curve (AUC). The AUC value is computed using the output score from the sigmoid activation function in the last layer. The accuracy is the average of the number of correct predictions over ground truth.

\section{Results}
The baseline conjoint model we are considering is a linear model defined in Eqn \ref{equ:conjoint} without the inclusion of additional hand-engineered interaction features. Conjoint usually is applied to estimate the impact of each attribute rather than predicting human decisions. You might notice that the baseline performance is rather low compared to other classification problem such as object recognition. This is partly due to that predicting human choices is inherently challenging as human decision process is noisy and complex. We can make completely different decisions with the same scenario under different circumstances.

\textbf{\begin{table}[htp]
\centering
\begin{tabular}{lcc}
\toprule
Model Type  & Accuracy & AUC \\
\midrule
Conjoint       &  0.719 & 0.779  \\
ConjointNet   & 0.789  &  0.850     \\
\bottomrule
\end{tabular}
\caption{Comparison on MM dataset with traditional conjoint analysis and the proposed ConjointNet with semi-supervised learning.}
\label{tab:mm}
\end{table}}

\subsection{Performance on the MM Dataset}
In this section, we evaluate the performance of our semi-supervised ConjointNet on the MM dataset. When using autoencoders, questions to ask are whether decoding input data through the latent representations can obtain successful reconstruction, and the number of nodes required at the bottleneck layer. We implemented two types of autoencoders, namely VAE and a plain autoencoder. The architecture of the encoder network consists of two fully connected layers, 276-dimensional input layer followed by a 128-dimensional hidden layer with batch normalizations and ReLU activations that project the input to a 2-dimensional latent space. The decoder network comprises two fully connected layers, 128-dimensional hidden layer that decode the projected 2-dimensional vectors, followed by a 276-dimensional output layer. 

Fig. \ref{fig:mm_recon} illustrates the obtained results that reconstruct from a testing input sample (left) with AE (right) and VAE (center) trained on 70\% of the MM data. Both AE and VAE show successful reconstruction with a 2-D latent space. This shows that the autoencoders generated representations that capture the input distribution in a reduced dimension, thus removing noise. It is easy to see that AE gives a better reconstruction compared to VAE on unseen data. This is not expected as AE is more prone to overfitting. One possibility is that the distributions in the MM dataset are relatively simple, a high hidden dimensionality without an increase in data complexity cause VAE to overfit the training dataset and learn unrepresentative features.

After the autoencoder is trained on raw input data without supervision. The second stage is to train the choice-classification network with human decision responses. The trained encoder (the first half of the AE in Fig. \ref{fig:ssl}) is connected to a two-layered fully connected classification network. We can train the netwrok in two ways: by freezing the encoder and only update the weights of the classification network or fine-tune the encode while the classification network is optimized. Table \ref{tab:mm} reports the choice prediction performance with our proposed model and the baseline conjoint on the MM dataset. The proposed ConjointNet neural network model outperforms the conjoint analysis by 7\% in both classification accuracy and AUC value. 

\begin{figure}[t]
    \centering
    \includegraphics[width=.9\linewidth]{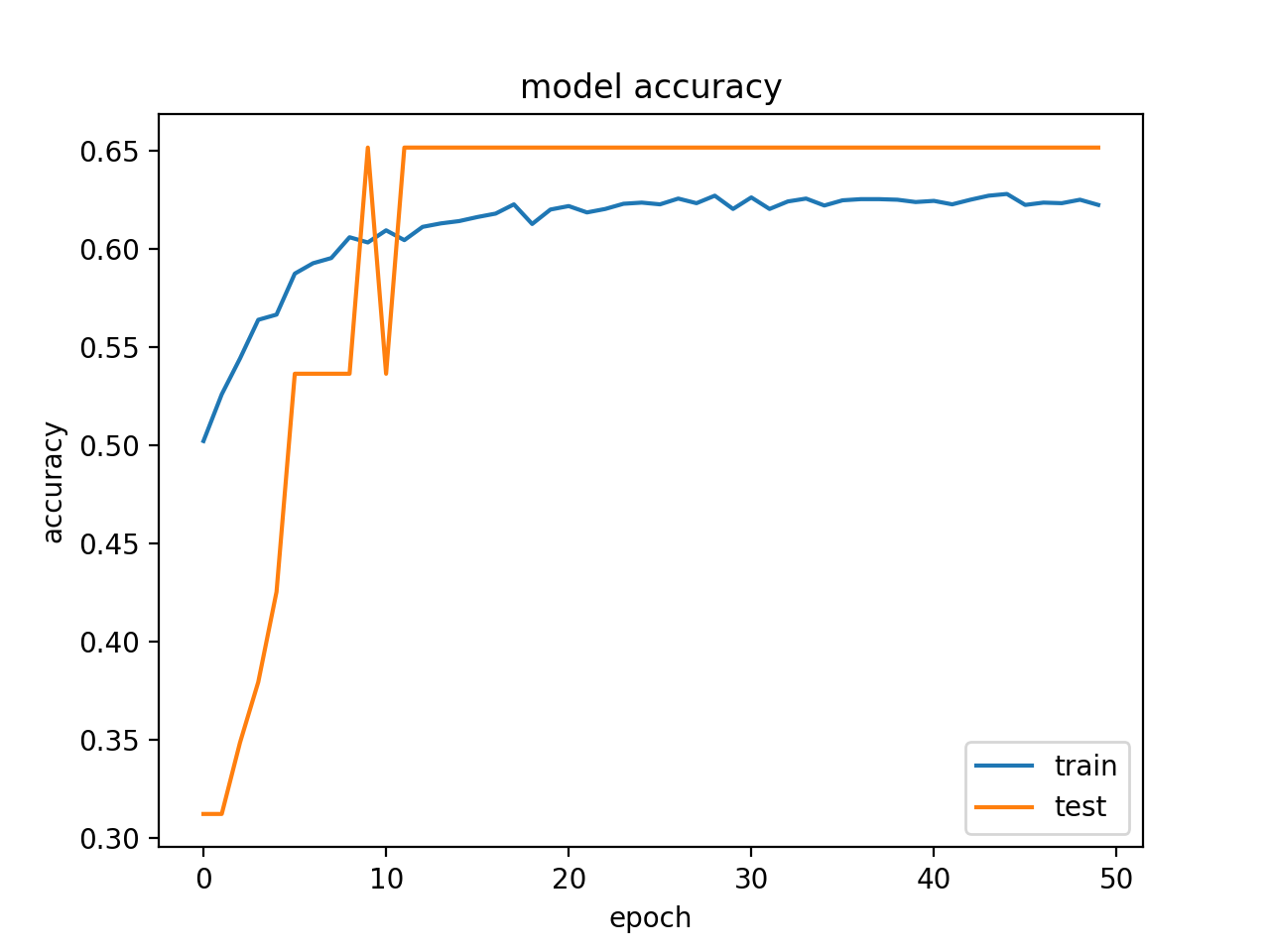}
    \caption{Training and testing accuracy on the car preference dataset with 16 hidden nodes.}
    \label{fig:car_16nodes}
\end{figure}

\textbf{\begin{figure}[t]
    \centering
    \includegraphics[width=.9\linewidth]{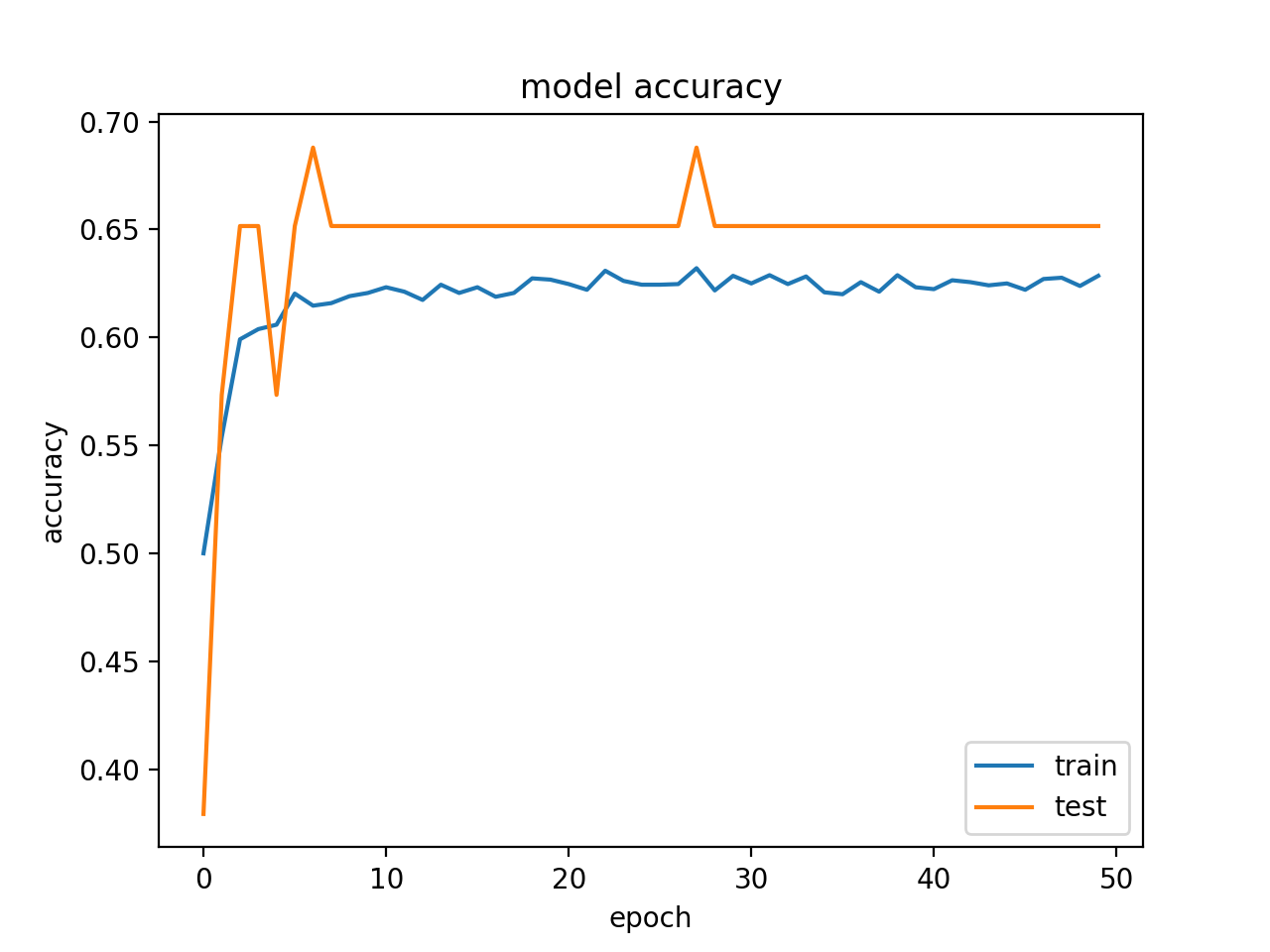}
    \caption{Training and testing accuracy on the car preference dataset with 64 hidden nodes.}
    \label{fig:car_64nodes}
\end{figure}}

\textbf{\begin{table}[t]
\centering
\begin{tabular}{lcc}
\toprule
Model Type & Accuracy & AUC \\
\midrule
Conjoint       & 0.618  & 0.659      \\
ConjointNet   &  0.688 & 0.661      \\
\bottomrule
\end{tabular}
\caption{Comparison on car preference dataset with traditional conjoint analysis and our proposed residual ConjointNet.}
\label{tab:car}
\end{table}}

\subsection{Performance on Car Preference}

We evaluated the performance of our residual ConjointNet model on the car preference dataset. Table \ref{tab:car} shows that ConjointNet improves the prediction accuracy by 7\% compared to traditional conjoint analysis. ConjointNet achieved similar AUC values as conjoint but a much higher prediction accuracy. Similar AUC values indicate that the two models have comparable performance when looking at all possible classification thresholds. When selecting an appropriate cutoff for the classification threshold, Conjoint can provide much higher prediction accuracy.

Fig. \ref{fig:car_16nodes} and Fig. \ref{fig:car_64nodes} illustrate the choice prediction results with 16 and 64 nodes in the hidden layer. As the number of nodes increases, the network has larger capacity, which leads to performance increasing by over 2\% in both training and testing accuracies. The improvement in training accuracy indicates that an increase in model capacity allows ConjointNet to extract representative features from the car preference data that are not possible with linear models. The increase in the testing accuracy with a higher number of nodes shows that our model did not overfit the training dataset. The complexity of the car preference data benefits from our proposed model compared to traditional simple linear models.

\section{Conclusion}
This paper proposed ConjointNet, two novel neural network architectures to predict user preferences. ConjointNet leverages representation learning to overcome the limitations of linear assumptions in traditional conjoint analysis. We demonstrated that ConjointNet outperforms conjoint in predicting user preferences on two public datasets. In particular, we observed that our model is not only suitable for data that were not designed explicitly for conjoint analysis (MM) but also outperforms conventional methods. This is a promising demonstration of ConjointNet's efficacy in more realistic settings. Future work will focus on the needs of the end user in these more realistic settings, investigating approaches for presenting and visualizing the non-linear feature interactions discovered by ConjointNet to end users.

\bibliographystyle{named}
\bibliography{ijcai22}

\begin{thebibliography}{}

\bibitem[\protect\citeauthoryear{Abbasnejad \bgroup \em et al.\egroup
  }{2013}]{Abbasnejad2013-jx}
Ehsan Abbasnejad, Scott Sanner, Edwin~V Bonilla, and Pascal Poupart.
\newblock Learning community-based preferences via dirichlet process mixtures
  of gaussian processes.
\newblock In {\em Proceedings of the {Twenty-Third} international joint
  conference on Artificial Intelligence}, IJCAI '13, pages 1213--1219. AAAI
  Press, August 2013.

\bibitem[\protect\citeauthoryear{Agrawal \bgroup \em et al.\egroup
  }{2020}]{Agrawal2020-cy}
Mayank Agrawal, Joshua~C Peterson, and Thomas~L Griffiths.
\newblock Scaling up psychology via scientific regret minimization.
\newblock {\em Proc. Natl. Acad. Sci. U. S. A.}, 117(16):8825--8835, April
  2020.

\bibitem[\protect\citeauthoryear{Amodei \bgroup \em et al.\egroup
  }{2016}]{amodei2016deep}
Dario Amodei, Sundaram Ananthanarayanan, Rishita Anubhai, Jingliang Bai, Eric
  Battenberg, Carl Case, Jared Casper, Bryan Catanzaro, Qiang Cheng, Guoliang
  Chen, et~al.
\newblock Deep speech 2: End-to-end speech recognition in english and mandarin.
\newblock In {\em International conference on machine learning}, pages
  173--182. PMLR, 2016.

\bibitem[\protect\citeauthoryear{Awad \bgroup \em et al.\egroup
  }{2018}]{Awad2018-zc}
Edmond Awad, Sohan Dsouza, Richard Kim, Jonathan Schulz, Joseph Henrich, Azim
  Shariff, Jean-Fran{\c c}ois Bonnefon, and Iyad Rahwan.
\newblock The moral machine experiment.
\newblock {\em Nature}, 563(7729):59--64, November 2018.

\bibitem[\protect\citeauthoryear{Bengio \bgroup \em et al.\egroup
  }{2013}]{Bengio2013-yx}
Yoshua Bengio, Aaron Courville, and Pascal Vincent.
\newblock Representation learning: a review and new perspectives.
\newblock {\em IEEE Trans. Pattern Anal. Mach. Intell.}, 35(8):1798--1828,
  August 2013.

\bibitem[\protect\citeauthoryear{Chapelle and
  Harchaoui}{2005}]{chapelle2005machine}
Olivier Chapelle and Za{\i}d Harchaoui.
\newblock A machine learning approach to conjoint analysis.
\newblock {\em Advances in neural information processing systems}, 17:257--264,
  2005.

\bibitem[\protect\citeauthoryear{Devlin \bgroup \em et al.\egroup
  }{2018}]{Devlin2018-ck}
Jacob Devlin, Ming-Wei Chang, Kenton Lee, and Kristina Toutanova.
\newblock {BERT}: Pre-training of deep bidirectional transformers for language
  understanding.
\newblock October 2018.

\bibitem[\protect\citeauthoryear{Green \bgroup \em et al.\egroup
  }{2001}]{green2001thirty}
Paul~E Green, Abba~M Krieger, and Yoram Wind.
\newblock Thirty years of conjoint analysis: Reflections and prospects.
\newblock {\em Interfaces}, 31(3\_supplement):S56--S73, 2001.

\bibitem[\protect\citeauthoryear{He \bgroup \em et al.\egroup
  }{2015}]{He2015-ms}
Kaiming He, Xiangyu Zhang, Shaoqing Ren, and Jian Sun.
\newblock Deep residual learning for image recognition.
\newblock December 2015.

\bibitem[\protect\citeauthoryear{Ho}{1995}]{Ho}
Tin~Kam Ho.
\newblock Random decision forests.
\newblock 1:278--282 vol.1, August 1995.

\bibitem[\protect\citeauthoryear{Kingma and Welling}{2013}]{Kingma2013-wu}
Diederik~P Kingma and Max Welling.
\newblock {Auto-Encoding} variational bayes.
\newblock December 2013.

\bibitem[\protect\citeauthoryear{Lai \bgroup \em et al.\egroup
  }{2015}]{Lai2015-qt}
Siwei Lai, Liheng Xu, Kang Liu, and Jun Zhao.
\newblock Recurrent convolutional neural networks for text classification.
\newblock In {\em Proceedings of the {Twenty-Ninth} {AAAI} Conference on
  Artificial Intelligence}, AAAI'15, pages 2267--2273. AAAI Press, January
  2015.

\bibitem[\protect\citeauthoryear{Loreggia \bgroup \em et al.\egroup
  }{2019}]{loreggia2019cpm}
Andrea Loreggia, Nicholas Mattei, Francesca Rossi, and K~Brent Venable.
\newblock Cpm etric: Deep siamese networks for metric learning on structured
  preferences.
\newblock In {\em International Joint Conference on Artificial Intelligence},
  pages 217--234. Springer, 2019.

\bibitem[\protect\citeauthoryear{O'Hara and
  Sillanp{\"a}{\"a}}{2009}]{o2009review}
Robert~B O'Hara and Mikko~J Sillanp{\"a}{\"a}.
\newblock A review of bayesian variable selection methods: what, how and which.
\newblock {\em Bayesian analysis}, 4(1):85--117, 2009.

\bibitem[\protect\citeauthoryear{Orme \bgroup \em et al.\egroup
  }{1997}]{orme1997assessing}
Bryan~K Orme, Mark~I Alpert, and Ethan Christensen.
\newblock Assessing the validity of conjoint analysis--continued.
\newblock In {\em Sawtooth Software Conference Proceedings}, pages 209--226.
  Citeseer, 1997.

\bibitem[\protect\citeauthoryear{Orme}{2006}]{orme2006getting}
Bryan~K Orme.
\newblock Getting started with conjoint analysis: strategies for product design
  and pricing research.
\newblock 2006.

\bibitem[\protect\citeauthoryear{Pfannschmidt \bgroup \em et al.\egroup
  }{2022}]{pfannschmidt2022learning}
Karlson Pfannschmidt, Pritha Gupta, Bj{\"o}rn Haddenhorst, and Eyke
  H{\"u}llermeier.
\newblock Learning context-dependent choice functions.
\newblock {\em International Journal of Approximate Reasoning}, 140:116--155,
  2022.

\bibitem[\protect\citeauthoryear{Radford \bgroup \em et al.\egroup
  }{2018}]{radford2018improving}
Alec Radford, Karthik Narasimhan, Tim Salimans, and Ilya Sutskever.
\newblock Improving language understanding by generative pre-training.
\newblock 2018.

\bibitem[\protect\citeauthoryear{Steiner and
  Mei{\ss}ner}{2018}]{steiner2018user}
Michael Steiner and Martin Mei{\ss}ner.
\newblock A user’s guide to the galaxy of conjoint analysis and compositional
  preference measurement.
\newblock {\em Marketing ZFP}, 40(2):3--25, 2018.

\bibitem[\protect\citeauthoryear{Steiner \bgroup \em et al.\egroup
  }{2016}]{steiner2016customer}
Michael Steiner, Roland Helm, and Verena H{\"u}ttl-Maack.
\newblock A customer-based approach for selecting attributes and levels for
  preference measurement and new product development.
\newblock {\em International Journal of Product Development}, 21(4):233--266,
  2016.

\bibitem[\protect\citeauthoryear{Tibshirani}{1996}]{Tibshirani1996-id}
Robert Tibshirani.
\newblock Regression shrinkage and selection via the lasso.
\newblock {\em J. R. Stat. Soc. Series B Stat. Methodol.}, 58(1):267--288,
  1996.

\bibitem[\protect\citeauthoryear{Toubia \bgroup \em et al.\egroup
  }{2007}]{toubia2007optimization}
Olivier Toubia, Theodoros Evgeniou, John Hauser, et~al.
\newblock Optimization-based and machine-learning methods for conjoint
  analysis: Estimation and question design.
\newblock {\em Conjoint measurement: Methods and applications}, 12:231--258,
  2007.

\bibitem[\protect\citeauthoryear{Unler \bgroup \em et al.\egroup
  }{2011}]{unler2011mr2pso}
Alper Unler, Alper Murat, and Ratna~Babu Chinnam.
\newblock mr2pso: A maximum relevance minimum redundancy feature selection
  method based on swarm intelligence for support vector machine classification.
\newblock {\em Information Sciences}, 181(20):4625--4641, 2011.

\end{thebibliography}

\end{document}